# Controlling Synthetic Characters in Simulations: A Case for Cognitive Architectures and Sigma

**Volkan Ustun, Paul S. Rosenbloom**
**Institute for Creative Technologies, USC**
**Playa Vista, CA**
**ustun@ict.usc.edu, rosenbloom@usc.edu**

**Seyed Sajjadi**
**Cal State University Northridge**
**Northridge, CA**
**seyed.sajjadi.947@my.csun.edu**

**Jeremy Nuttall**
**Univ. of Southern California**
**Los Angeles, CA**
**jeremy.nuttall@usc.edu**

## ABSTRACT

Simulations, along with other similar applications like virtual worlds and video games, require computational models of intelligence that generate realistic and credible behavior for the participating synthetic characters. Cognitive architectures, which are models of the fixed structure underlying intelligent behavior in both natural and artificial systems, provide a conceptually valid common basis, as evidenced by the current efforts towards a *standard model* of the mind, to generate human-like intelligent behavior for these synthetic characters. Developments in the field of artificial intelligence, mainly in probabilistic graphical models and neural networks, open up new opportunities for cognitive architectures to make the synthetic characters more autonomous and to enrich their behavior. Sigma (Σ) is a cognitive architecture and system that strives to combine what has been learned from four decades of independent work on symbolic cognitive architectures, probabilistic graphical models, and more recently neural models, under its *graphical architecture hypothesis.* Sigma leverages an extended form of factor graphs towards a uniform *grand unification* of not only traditional cognitive capabilities but also key non-cognitive aspects, creating unique opportunities for the construction of new kinds of cognitive models that possess a *Theory-of-Mind* and that are perceptual, autonomous, interactive, affective, and adaptive. In this paper, we will introduce Sigma along with its diverse capabilities and then use three distinct proof-of-concept Sigma models to highlight combinations of these capabilities: (1) Distributional reinforcement learning models in a simple OpenAI Gym problem; (2) A pair of adaptive and interactive agent models that demonstrate rule-based, probabilistic, and social reasoning in a physical security scenario instantiated within the SmartBody character animation platform; and (3) A knowledge-free exploration model in which an agent leverages only architectural appraisal variables, namely attention and curiosity, to locate an item while building up a map in a Unity environment.

## ABOUT THE AUTHORS

**Volkan Ustun** is a senior artificial intelligence researcher at the USC Institute for Creative Technologies. His general research interests are cognitive architectures, computational cognitive models, natural language and simulation. He is a member of the cognitive architecture group and a major contributor to the Sigma cognitive architecture.

**Paul S. Rosenbloom** is a professor of computer science at the USC Viterbi School of Engineering and director for cognitive architecture research at the USC Institute for Creative Technologies. His general research interests are cognitive architectures (all aspects), a Common Model of Cognition (née a Standard Model of the Mind), and computing as a great scientific domain. He was one of the co-developers of the Soar cognitive architecture and the initial developer of the Sigma cognitive architecture.

**Seyed Sajjadi** is a machine learning researcher at Hughes Research Laboratories (HRL) and a volunteer member of the cognitive architecture group at USC ICT. His general research interest lies primarily in the theory and application of artificial intelligence, particularly in developing novel approaches for interactive machine learning and structured prediction.

**Jeremy Nuttall** is a fifth-year student in Computer Science at the University of Southern California. His interests include cognitive architectures, philosophy of mind, and the intersection between AI, neuroscience, and philosophy of mind.

# Controlling Synthetic Characters in Simulations: A Case for Cognitive Architectures and Sigma

**Volkan Ustun, Paul S. Rosenbloom**
**Institute for Creative Technologies, USC**
**Playa Vista, CA**
**ustun@ict.usc.edu, rosenbloom@usc.edu**

**Seyed Sajjadi**
**Cal State University Northridge**
**Northridge, CA**
**seyed.sajjadi.947@my.csun.edu**

**Jeremy Nuttall**
**Univ. of Southern California**
**Los Angeles, CA**
**jeremy.nuttall@usc.edu**

## INTRODUCTION

More than twenty years ago, Tambe et al. (1995) discussed the generation of human-like synthetic characters that can interact with each other, as well as with humans, within the emerging domain of highly interactive simulations. Twenty years later, developments in computer graphics and animation have allowed for extremely realistic-looking interactive simulation environments; it is now possible to create almost photo-real synthetic characters with realistic gaits and gestures. However, progress in behavior generation has been more mixed. Today, many interactive simulation models desire synthetic characters that are not only believable but also indistinguishable from humans – performing coherent sequences of believable behavior sustained across different tasks and environments. These characters are almost impossible to generate by strictly controlling their behavior, as this approach would require generating rules for each different task, each different environment, and every possible interaction.

Consequently, generating such behavior is still considered a challenge by/for the simulation community (Taylor et al. 2015), especially when there is a need for creating human-like autonomous social cognitive (HASC) systems that have general perception, action, learning, and social capabilities appropriate for a variety of environments. One of the most prominent approaches to tackling this challenge is utilizing *cognitive architectures* (Langley, Laird & Rogers, 2009) that specify the fixed structures and processes of cognition that should be approximately the same across all human-like agents while remaining constant across the lifetime of each such agent. This includes memory systems for agent's beliefs, goals, and experience; knowledge representation; functional operations that lead from perception through behavior; and learning mechanisms (Newell 1990). A cognitive architecture by itself does not yield behavior, but when supplemented by more variable knowledge and skills it yields a cognitive system (or a cognitive agent) that serves as a computational model of intelligence that can reason and participate in a variety of simulations, virtual worlds, and virtual games.

In this paper, first, a case is made for utilizing cognitive architectures in cognitive behavior model development rather than devising ad-hoc models with narrow scope for behavior generation. The *Sigma* cognitive architecture is a relatively recent architecture with a unique perspective and a unique set of capabilities (Rosenbloom, Demski & Ustun, 2016a). The rest of this paper focuses on why Sigma is one of the most promising approaches to model HASC systems in interactive simulation environments, while leveraging three distinct proof-of-concept models as evidence for this argument.

### Why Cognitive Architectures?

Utilizing cognitive architectures in the model development process could significantly benefit four major aspects of the development of computational behavior models for synthetic characters:

1) *Leverage* - Both the fixed processes and the tuned parameters of cognition that are relatively invariant across time and domains, and that are essential for human-like behavior, can be leveraged in a cognitive architecture to model a range of requisite cognitive phenomena. Such reuse across a variety of complex task models provides a critical advantage while building up cognitive systems (Salvucci 2006).
2) *Constrain* - Cognitive architectures impose constraints on how things can be modeled. These constraints frame modelers' thinking around the mechanisms and processes made available by the architecture while devising a cognitive system, rather than permitting or even encouraging an assembly of ad-hoc models narrowly scoped for a specific task. Such an endeavor helps to create cognitive systems encapsulating the primitives of cognition as envisioned in the design of the utilized cognitive architecture by impacting how one thinks about modeling a task to potentially lead to a deeper level of understanding of the cognitive phenomena underlying the task at hand (Sun 2007)

3) *Validate* - Validation of behavior models for human-like intelligent characters is a significant yet tricky task for interactive simulation models. Sargent (2011) defines two validation and verification mechanisms that apply to the conceptual part of the model development process: (1) *Conceptual model validation* and (2) *Computerized model verification*. Utilizing cognitive architectures that have themselves been previously validated can provide substantial evidence for the conceptual validity of the cognitive model at hand, with only the variable portion needing additional attention. Validation of cognitive architectures has recently been simplified by the development of a nascent community consensus concerning a Standard Model of the Mind (Laird, Lebiere & Rosenbloom, 2017), enabling validation of a sort by a mapping onto this consensus. The standard model focuses on many of the critical aspects of cognition including structure and processing, memory and content, learning, perception, and motor; and has been further supported by empirical evidence from neuroimaging data (Stocco et al., 2018). Furthermore, cognitive architectures most often provide their own unique language, syntax, and semantics to assist in generating cognitive systems based on the architecture, in guiding the implementation of the conceptual cognitive system model and, hence, in helping to verify the computerized (or implemented) conceptual cognitive system model.
4) *Integrate* - Human-like autonomous social cognitive (HASC) systems must at least be able to: behave on their own via perception, reasoning, and action; use a Theory of Mind while interacting and communicating naturally with others; be affective, and motivated; and learn pervasively from their experience (Swartout, 2010). Additionally, these capabilities must be integrated together and work coherently for complete HASC systems. This integration can be quite challenging, yet can potentially yield much more than the sum of its parts. Cognitive architectures strive to maintain a fully working cognitive system at all times and mechanisms for coherently integrating the listed capabilities are integral to their design, whereas ad-hoc combinations of narrowly scoped behavior models make such integration particularly challenging.

Cognitive architectures have controlled virtual characters capably in many interactive simulation environments. Mainstream cognitive architectures, including Soar (Laird, 2012) and ACT-R (Anderson, 2007) with more than four decades worth of development behind them, originated as production systems and are reasonably capable of modeling the reactive, knowledge-intensive, and goal-driven aspects of human behavior. For example, the Soar cognitive architecture was used to model the behavior of pilots and commanders for fixed-wing (Jones et al., 1999) and rotary-wing (Hill et al., 1997) aircraft in complex entity-level military simulations, whereas ACT-R was utilized to model a taxiing commercial jetliner (Zemla et al., 2011). Hill et al. (2003) integrated a Soar-based intelligent system with a virtual reality training environment for the Mission Rehearsal Exercise, an experiential learning system designed to teach critical decision-making skills to small-unit leaders in the U.S. Army. Similarly built characters were leveraged in a single party (Traum et al., 2005) and multi-party (Kim et al., 2009) negotiation training exercises in cross-cultural contexts. Another relevant example was the utilization of Soar to model a synthetic character with multiple goals and complex tactics in the computer game Quake II (Laird & VanLent, 2001). Although these models all showed potential to create realistic and believable synthetic characters, none of them embodied probabilistic reasoning or learning, both of which are essential for adaptation to new environments and to the noise in current environments as evidenced in many successful machine learning practices of today.

## Why Sigma?

As Swartout (2010) pointed out, behaving like real people requires HASC systems to, among other things: (1) use their perceptual capabilities to observe their environment and other synthetic characters and humans in it; (2) act autonomously in their environment based on what they know and perceive, e.g. reacting and appropriately responding to the events around them; (3) interact in a natural way with both real and other virtual humans using verbal and nonverbal communication; (4) possess a Theory of Mind (ToM) to model their own mind and the minds of others; (5) understand and exhibit appropriate emotions and associated behaviors; and (6) adapt their behavior through experience. The Soar and ACT-R communities worked toward addressing these six capabilities for synthetic characters, but some items were just not feasible within the core architecture. For example, external modules were required for acceptable perceptual and communication capabilities. Likewise, most of the emotion models were also outside the core. More importantly, they have not been able to fully capture the advances that have been made in recent years in behavioral adaptation, or in other words, learning. Some aspects of learning were successfully incorporated, but generality in statistical machine learning has eluded them. For example, probabilistic graphical models (Koller & Friedman, 2009) provide a general tool that combines graph theory and probability theory to enable efficient probabilistic reasoning and learning in ways that haven't been possible with cognitive architectures based on production systems.

Similarly, recent developments in deep neural models (Goodfellow, Bengio & Courville, 2016) do not have a direct reflection in traditional cognitive architectures. The machine learning community employs such models as one of its primary tools, yielding state of the art results for at least four of the listed capabilities that have challenged traditional cognitive architectures: perception, autonomy, interaction, and adaptation. However, most of these improvements have been achieved independently, with little effort toward integration across them, although a few neural network models are pushing towards being more comprehensive in some limited capacity (Graves, Wayne & Danihelka, 2014; Weston, Chopra & Bordes 2015; Sukhbaatar, Weston & Fergus, 2015). For any HASC system, on the other hand, these capabilities need to be integrated together and work coherently.

Sigma (Σ) (Rosenbloom, Demski & Ustun, 2016a), which is actively developed and maintained at the USC Institute for Creative Technologies (ICT), is a cognitive architecture and system that starts from a theoretically elegant yet broadly applicable and efficient hybrid (discrete + continuous) mixed (symbolic + probabilistic) base, grounded in probabilistic graphical models, with a long-term objective of understanding and replicating the architecture of a mind in a manner that is: (1) *grand unified*, spanning not only traditional cognitive capabilities but also crucial sub-cognitive aspects such as perception, motor control, and affect; (2) *generically cognitive*, capturing the essence of both natural and artificial intelligence at a suitable level of abstraction; (3) *functionally elegant*, yielding broad cognitive and sub-cognitive functionality – ultimately all that is necessary for human-like intelligence – from a simple and theoretically elegant base; and (4) *sufficiently efficient*, executing quickly enough for large-scale experiments in modeling human cognition and for real-time applications involving virtual humans, intelligent agents or robots.

Sigma is based on retaining what still seems to be right about Soar by combining lessons from it as a symbolic cognitive architecture with probabilistic graphical models, and more recently neural models, under its graphical architecture hypothesis (Rosenbloom, Demski & Ustun, 2016a). In particular, Sigma leverages a generalization of factor graphs (Kschischang, Frey & Loeliger, 2001) towards a uniform grand unification of not only traditional cognitive capabilities but also of critical non-cognitive aspects, creating unique opportunities for the construction of new kinds of cognitive models that possess a Theory of Mind and that are perceptual, autonomous, interactive, affective, and adaptive. Sigma's graphical architecture has even recently been extended to handle neural networks (Rosenbloom, Demski & Ustun, 2016b; Rosenbloom, Demski & Ustun, 2017). Sigma has quite general parameter-learning capabilities, in that probabilistic, neural, and reinforcement learning all emerge from a local gradient-descent-based learning mechanism operating at the core of the architecture (Rosenbloom, 2012a; Rosenbloom et al., 2013; Rosenbloom, 2014; Rosenbloom, Demski & Ustun, 2017). In summary, Sigma is a well-equipped candidate to tackle the challenge of modeling adaptive synthetic characters as HASC systems for successful interactive simulations. Furthermore, the aspiration for grand unification plus its particular blend of capabilities put Sigma on a unique path towards achieving human-like intelligence for HASC systems, with it being the first, and still only, attempt to do so via a deep fusion of graphical models, neural networks, and cognitive architectures.

The next section provides a brief introduction to the Sigma cognitive language. Then, Sigma's potential for creating HASC systems will be discussed along with three disparate proof-of-concept Sigma models – a basic RL model combined with symbolic rules, a pair of ambulatory agents in a physical security setting (Ustun & Rosenbloom, 2015), and a knowledge-free exploration model leveraging only architectural appraisal variables– that demonstrate its diverse capabilities.

## THE SIGMA COGNITIVE LANGUAGE

The Sigma cognitive architecture is built on *factor graphs* (Kschischang, Frey & Loeliger, 2001) – undirected graphical models (Koller & Friedman, 2009) with variable and factor nodes, and functions that are stored in the factor nodes. Graphical models provide a general computational technique for efficient computation with complex multivariate functions – implemented via hybrid mixed *piecewise-linear functions* in Sigma (Rosenbloom, Demski & Ustun, 2016a) – by leveraging forms of independence to: decompose them into products of simpler functions; map these products onto graphs; and solve the graphs via message passing or sampling methods. The *summary product algorithm* (Kschischang, Frey & Loeliger, 2001) is the general inference algorithm in Sigma (Figure 1).

Factor graphs are particularly attractive as a basis for broadly functional, yet simple and theoretically elegant, cognitive architectures because they are not limited to just probabilistic reasoning, instead providing a single general representation and inference algorithm for processing symbols, probabilities, distributed neural vectors, and signals.

Sigma defines a high-level language of *predicates* and *conditionals* that compiles down into factor graphs. Predicates specify relations over continuous, discrete and/or symbolic arguments. They are defined via a name and a set of typed arguments, with *working memory* (WM) containing predicate instantiations as functions within a WM sub-graph. Predicates may also have perception and/or long-term memory (LTM) functions. For perceptual predicates, factor nodes for *perceptual buffers* are connected to the corresponding WM sub-graphs. For example, `Observed(object:object visible:boolean)` is a perceptual predicate with two arguments: (1) `object` of type `object`; and (2) `visible` of type `boolean`. This predicate specifies which objects are visible to the agent at any particular time. For memorial predicates, *function factor nodes* (FFNs) are likewise connected to the corresponding WM sub-graphs. Messages into FFNs provide the gradients for learning the nodes' functions. Gradient calculations require identifying parent-child relationships among predicate arguments; for instance, the predicate function for the `Object-Location-X(object:object x:location%)` predicate defines a distribution over the `x` coordinate given an object (`x` is marked as the child variable in this predicate by `%`).

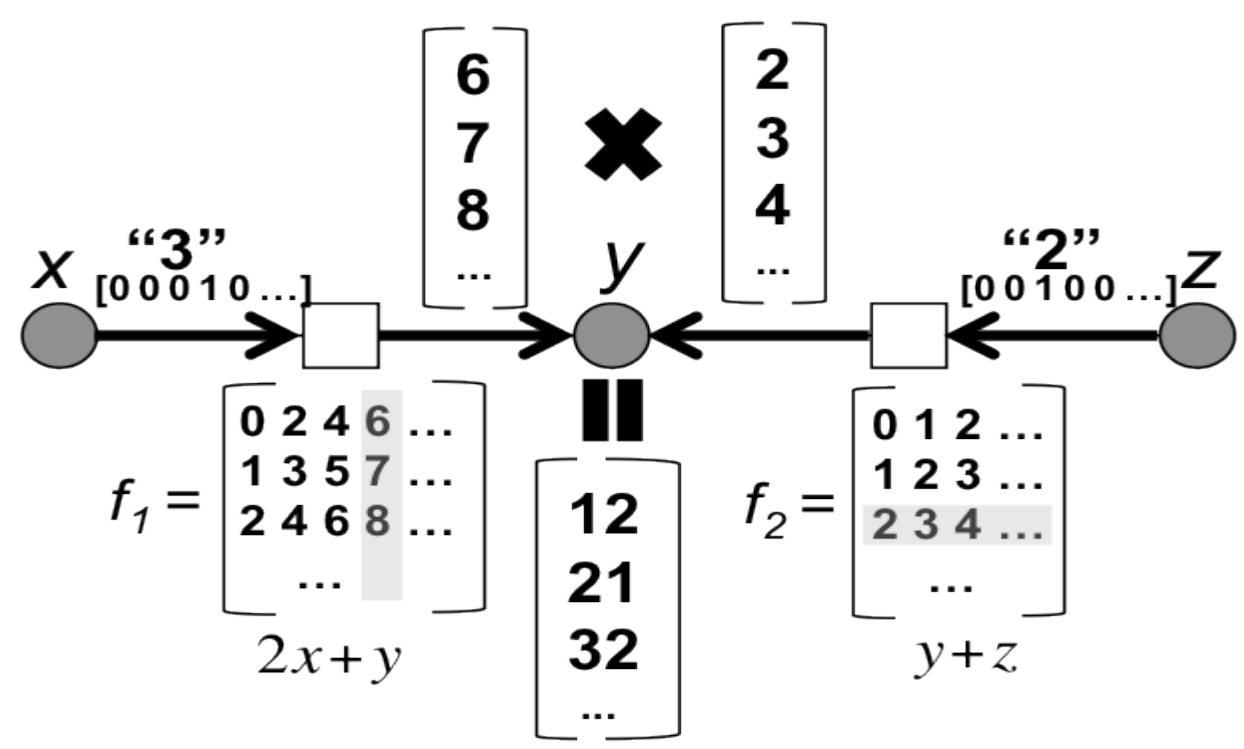

**Figure 1.** Summary product computation over the factor graph for $f(x,y,z) = y^2+yz+2yx+2xz = (2x+y)(y+z) = f_1(x,y)f_2(y,z)$ of the marginal on $y$ given evidence concerning $x$ and $z$.

*Conditionals* structure LTM and basic reasoning, compiling into more extended sub-graphs that interconnect with the appropriate WM sub-graphs. Conditionals are defined via a set of *predicate patterns* in which type specifications are replaced by constants and variables, plus an optional *function* over pattern variables. *Conditions* and *actions* are predicate patterns that behave like the respective parts of rules, pushing information in one direction from the conditions to the actions. The example conditional in Figure 2 updates the information about which objects have been seen so far, based on the information in the `Observed` predicate (where *o* is a variable over objects). *Condacts* are predicate patterns that support the bidirectional processing that is key to probabilistic reasoning, partial matching, constraint satisfaction and signal processing. Examples of probabilistic networks based on condacts and functions will be seen in the coming sections on virtual humans. Overall, conditionals provide a deep blend of rule systems and probabilistic networks.

The core processing in Sigma is driven by a cognitive cycle that comprises input, graph solution, decisions, learning, and output. Graph solution yields inference via the summary product algorithm, by *product* of the messages coming into each node – including the node's function when it is a factor node – and then *summarization*, via integration or maximum, of unneeded variables from outgoing messages. Perception and action are modeled as sub-symbolic reasoning in Sigma that occurs predominantly within graph solution, rather than within external modules (Chen et al., 2011; Joshi, Rosenbloom & Ustun, 2014; Joshi, Rosenbloom & Ustun, 2016; Rosenbloom, Demski & Ustun, 2017). Specifically, perceptual information collected during the input phase is passed to WM, where it is combined with the current contents of the WM and knowledge available within the LTM – such as might define a hidden Markov model (HMM), a conditional random field (CRF), or a neural network – yielding new WM content at the end of the graph solution phase.

```
CONDITIONAL Seen
   Conditions: Observed(object:o visible:true)
   Actions: Seen-Objects(object:o)
```

**Figure 2.** Conditional for context information.

Decisions in Sigma, in the classical sense of choosing one among the best operators to execute next, are based on an architecturally distinguished selection predicate – `Selected(state:state operator:operator!)` – with typically the operator associated with the highest value, or utility, in the distribution for a state being selected (as marked by "!"). The learning phase then modifies the long-term functions in the graph via a general form of gradient-descent learning. Finally, in the output phase, selected actions are performed in the outside world.

A single cognitive cycle yields *reactive processing*; a sequence of them yields *deliberative processing*; and *reflective processing* occurs when no operator can be selected, or a selected operator can't be applied, similar to impasses in Soar (Laird, 2012). Activities such as classifying an object or executing a body of rules can all occur reactively, within a single cognitive cycle. Deliberative processing utilizes reactive processing, and thus individual cognitive cycles, as its inner loop in providing sequential, algorithmic or knowledge-driven behavior that leads to problem solving and reasoning. When impasses occur, deliberative processing – either in the same or a different problem space – can be used reflectively to yield the knowledge that would resolve the impasses.

## SAMPLE SIGMA COGNITIVE MODELS

Three disparate proof-of-concept Sigma cognitive models, all involving embodied characters in virtual or game environments, are discussed in this section to highlight different capabilities and how they are integrated with Sigma. These models have mainly been developed as part of efforts to interface Sigma with different platforms, with each model demonstrating a different kind of integration. Not included here is work on other types of models, such as conversational agents (Ustun & Rosenbloom, 2016), that have not been developed as embodied characters.

The first proof of concept is the Frozen Lake model, which is a toy Reinforcement Learning (RL) problem available on the OpenAI Gym toolkit (https://gym.openai.com). The base model leverages Sigma's Reinforcement Learning template (Rosenbloom, 2012a), but augmented with domain knowledge in the form of Sigma rules that seamlessly interact with the RL model over the common factor graph representation.

The second proof of concept involves a pair of adaptive, interactive agents animated in a virtual environment for a typical retail store physical security scenario. This work was first discussed in (Ustun & Rosenbloom, 2015). Here, certain parts are reused with a few improvements in its presentation, as this work contributes to the main arguments made in this paper. The agents in this model are adaptive not only in dynamically deciding what to do based on the immediate circumstances but they also embody two distinct forms of relevant learning: (1) the automated acquisition of maps of the environment from experience with it, in the context of the classic robotic capability of Simultaneous Localization and Mapping (SLAM); and (2) reinforcement learning (RL) to improve decision making based on experience with the outcomes of earlier decisions. SmartBody (Shapiro, 2011) was used as the character animation platform for this study.

For the last proof of concept, a knowledge-free exploration model has been constructed. This model has showcased an agent leveraging only architectural appraisal variables – an idea adapted from computational emotion modeling (Scherer, Schorr & Johnstone, 2001; Moors et al., 2013) – namely *attention* and *curiosity*, to locate an item while building up a map in a (virtual) physical environment. This model is guided by a sense of curiosity that is itself grounded in a combination of familiarity and surprise, and it demonstrates the integration of architectural appraisal variables in a cognitive model controlling a synthetic character. The Unity game engine (https://unity3d.com/) has been used here as the animation platform.

All three models leverage VH Messages in the VHMsg library of the ICT Virtual Human Toolkit (2009) to establish communication between the Sigma model and the OpenAI, SmartBody, and Unity platforms.

### The Frozen Lake Model

The frozen lake model is a simple grid-based model, where an agent tries to receive a reward by reaching a goal location. Most of the grid cells are walkable, but some of the cells contain holes that the agent can fall in, effectively ending the current trial. The most compelling issue for this model is that there is significant uncertainty in the movement direction, a move can lead to – with equal probability –3 different neighboring grid locations. RL is one of the most standard techniques that can be leveraged here and Sigma has a standard RL template (Rosenbloom, 2012) that can be utilized in a straightforward manner for this.

The core idea for deriving an RL algorithm in Sigma has been to leverage gradient descent in learning of Q values over multiple trials, given appropriate conditionals to structure the computation as is needed for this to happen. The

`Compute-Backup-Value` conditional in Figure 3 examines the previous location (`location`), the current location (`location*next`), the local reward (`reward`), and the future discounted reward (`projected`). The `value` argument captures the utility in the range [-8, 12] for this model, where -8 represents the penalty of falling into a hole and +12 is the reward for reaching the goal location. In the actions, the conditional leverages an affine transformation, with an offset to add the current location's local reward to a `backup-value` distribution, and a coefficient to discount this sum. Separate conditionals then utilize the `backup-value` as a gradient to update the `projected` predicate function for the previous location and the `Q` predicate function for the previous location and the applied operator.

```
CONDITIONAL COMPUTE-BACKUP-VALUE
   Conditions: Location(x:x y:y)
               Location*Next(x:nx y:ny)
               Projected*Next(x:nx y:ny value:p)
               Reward(x:nx y:ny value:r)
   Actions: Backup-Value(x:x y:y value:.95*(p+r))
```

**Figure 3.** Conditional for computing the backup value.

The uncertainty in movements provides a significant challenge for the standard RL model in Sigma, with the agent having difficulty finding a path to the goal location for the frozen lake model. To rectify this issue, it is augmented with two forms of domain knowledge that are learned through experience: (1) an action model, and (2) the locations of the holes. The augmented model leverages this knowledge in its decision-making process map by estimating the potential of ending in a hole for a considered move and making the risky moves less likely to be selected as the experiment progresses. Drastic improvement is achieved in the model performance by this augmentation, as the average likelihood for reaching the goal location in the testing phase hit 79%. Here, the integration between the RL model and various forms of knowledge is facilitated by functional elegance, as the simple and elegant base of Sigma that leverages factor graphs, message passing, and gradient descent learning has made this integration seamless.

**The Physical Security System Model**

Physical security systems are comprised of structures, sensors, protocols, and policies that aim to protect fixed-site facilities against intrusions by external threats, as well as unauthorized acts by insiders. Physical security systems are generally easy to understand but they also allow complex interactions to emerge among the agents. A typical retail-store shoplifting plot, where offenders first pick up merchandise in a retail store and then try to leave without getting caught by any of the store's security measures, is a typical physical security system example. A simple grab-and-run scenario – the intruder needs to locate the desired item in the store, grab it, and then leave the store – is considered, although a large number of different scenarios are possible. In this scenario, the role of security is to detain the intruder before s/he leaves the store. A basic assumption is that it is not possible to tell what the intruder will do until s/he picks up an item and starts running. Security can immediately detect the activity and start pursuing the intruder once the item is picked up (assuming CCTV). If the intruder makes it to the door before security, it is considered a success for the intruder.

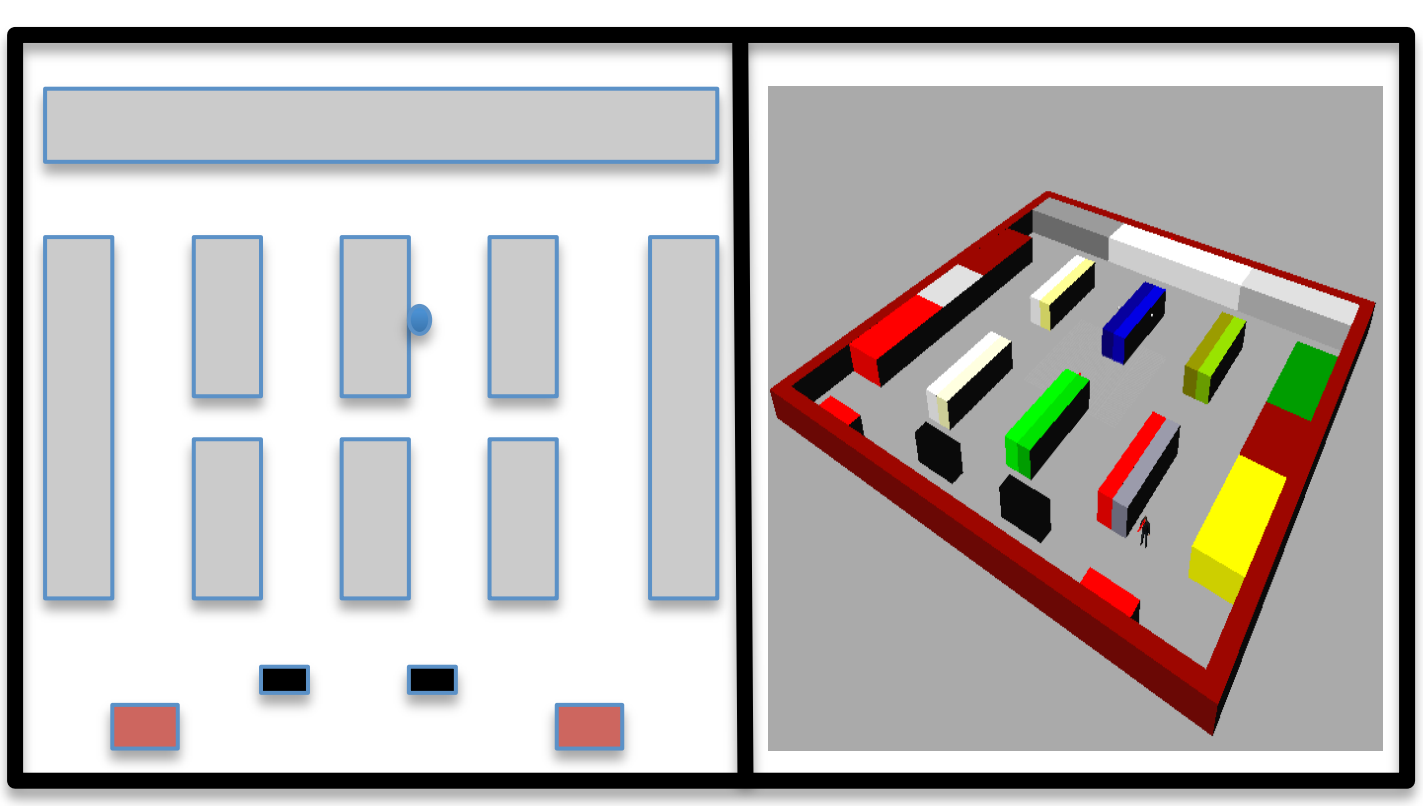

**Figure 4.** Layout of the store and its SmartBody representation**.**

For the basic setup, it is assumed that the intruder does not know the layout of the store and hence it has to learn a map and be able to use it to localize itself in the store. When the intruder locates the item of interest, it grabs the item and leaves the store via one of the exits. In the hypothetical retail store used here (Figure 4), there are shelves (gray rectangles), the item of interest (the blue circle) and two entry/exit doors (red rectangles). The intruder leaves the store via either (1) the door it used to enter or (2) the door closest to the item of interest. The main task for security is to learn about the exit strategies of intruders and use this to effectively detain them.

In this setup, locomotion and path finding are delegated to the SmartBody engine; Sigma sends commands and queries to SmartBody to perform these tasks and to return perceptual information. Two basic types of perception are utilized by the Sigma cognitive model: (1) information about the current location of the agent, mimicking the combination of direction, speed and odometry measurements available to a robot; and (2) objects that are in the visual field of the agent, along with their relative distances, mimicking the perception of the environment for a robot. Location information is conveyed to the Sigma cognitive model with noise added – perfect location information is not available to the model.

For this model, an intruder and a security agent are modeled as synthetic characters. There are two distinct types of learning (and probabilistic inference) in this scenario: (1) The intruder does not know the layout of the store in advance, and so it must learn a map of the store while simultaneously localizing itself in the learned map (SLAM model), and (2) The security agent infers the strategy of the intruder – i.e., whether it exits through the entry door or the closest door – by first learning a policy for the intruder via RL and then using this policy and the intruder's actions to determine on each trial the relative likelihoods of the two strategies being used. Here, the agents learn both the probability distributions in a Bayesian network and Q functions for an RL algorithm using the same basic set of architectural mechanisms, something that would require multiple distinct modules in other architectures if it could be done at all. The intruder also needs a decision framework to dynamically decide what to do based on its immediate circumstances.

In the SLAM model, the intruder has no a priori knowledge about the layout of the retail store (which is as shown in Figure 4). Therefore, it needs to learn a map of the store while simultaneously using the map to localize itself in the store. A 31×31 grid is imposed on the store for map learning. A virtual human only occupies a single grid cell, whereas objects in the environment – such as shelves – can span multiple cells. While performing SLAM, a Dynamic Bayesian Network (DBN) is utilized (Grisetti et al., 2010). In this representation (Figure 5), *l* is the location, *u* captures the odometry readings, *p* represents perception of the environment, and *m* is the map of the environment. The Sigma model defines two perceptual predicates – `Location-X(x:location)` and `Location-Z(z:location)` – to represent the location of the virtual human on the grid; here, the `location` type is discrete numeric, with a span of 31. Together these two predicates represent the space of 2D cells in the grid. They are perceptual predicates, and hence induce perceptual buffers that emulate odometry readings. In particular, the current location of the agent is perceived with noise – the default noise model assumes that any neighboring cell of the correct cell may be perceived as the agent's current location. In addition, the objects in the visual field are also perceived along with the relative distances of the center of these objects to the agent's location.

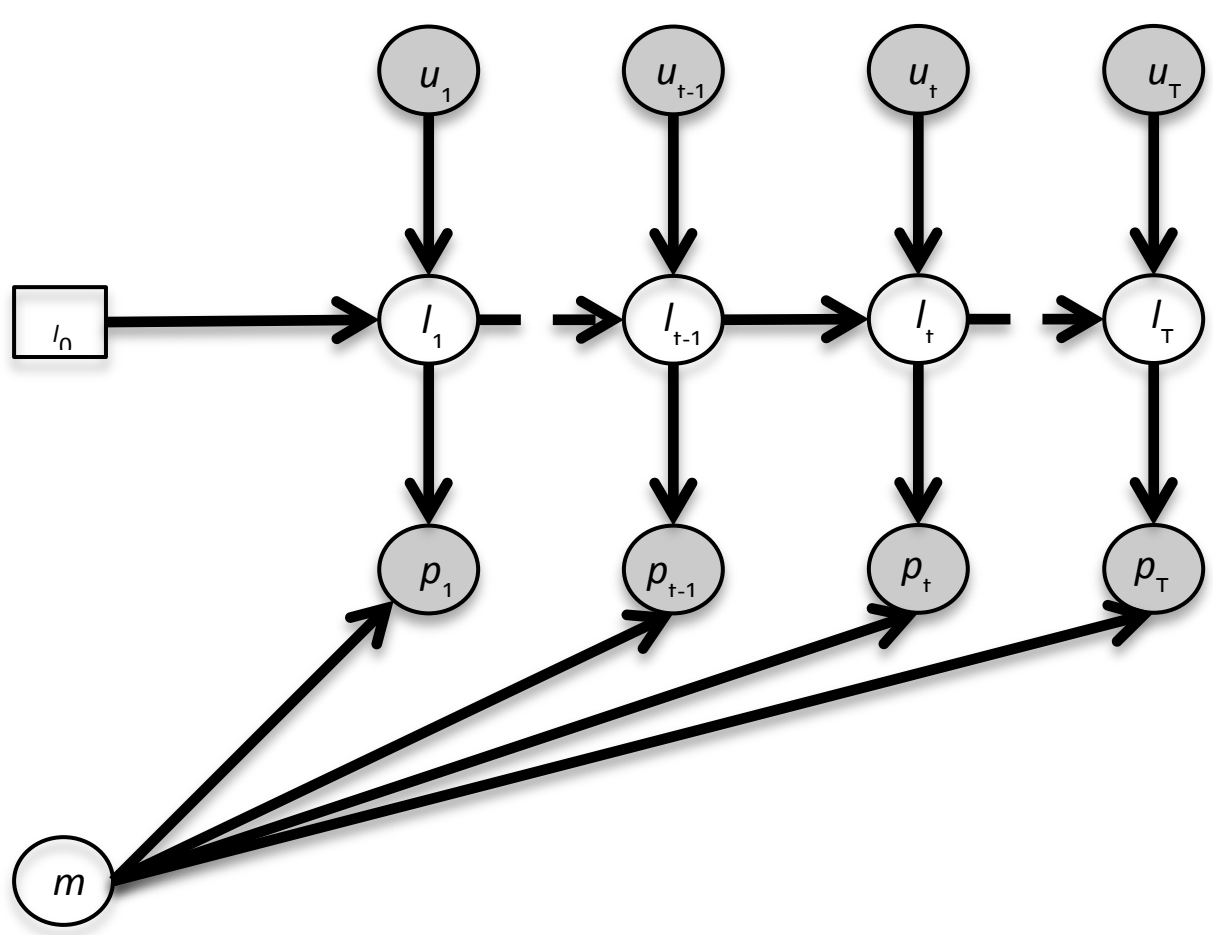

**Figure 5.** Dynamic Bayesian Network for the SLAM process.

The DBN is captured in two almost identical conditionals, one for `x` (Figure 6) and a similar one for `z`. These conditionals convert relative locations of objects given the agent to absolute locations in the map. They use Sigma's general capability for affine transforms in visual imagery (Rosenbloom, 2012b) to offset the agent's current location by the distance to the object being updated. In Figure 6, the `Object-Location-X` predicate is a memorial predicate, so it has a function that learns the x coordinates in the map via gradient descent. Since both the `Location-X` and `Object-Location-X` patterns are condacts, the processing is bidirectional between them; both perception of the VH's location and perception of the object locations have an impact on the posterior for the VH's location. This

```
CONDITIONAL SLAM-X
   Conditions: Observed(object:o visible:true)
               Object-Distance-X(dist-x:dx object:o)
   Condacts: Location-X(x:lx)
             Object-Location-X(object:o x:(lx-dx))
```

**Figure 6.** SLAM conditional for the *x* coordinate.

bidirectional processing forms the basis for SLAM, where the map is learned while it is simultaneously used for localization.

The security agent, in addition to mapping and localizing itself like the intruder, needs to reason about the intruder's actions for effective detainment, and hence it needs to learn about the policies used by the intruder. One basic assumption made here is that it is easy to recognize that a grab-and-go scenario has been initiated, by observing the pick-up behavior of the intruder. However, even though security can easily recognize when such a scenario has been initiated, it still needs to intercept the intruder before it leaves the retail store. As there are two exit doors, early anticipation of the intruder's choice increases the chances of a successful detention.

Theory of Mind (ToM) involves formation of models of others and generation of expectations about their behavior based on these models to enable effective decisions in social settings (Whiten, 1991). In decision theoretic approaches to ToM, such models can be represented as reward functions. For the intruder in our scenario, there are two possible models, distinguished by whether a reward is received when the agent returns to its door of entry or when it reaches the nearest door from the item of interest. This corresponds to Bayesian approaches to multi-agent modeling that use a distribution over a set of policies to specify the beliefs that one agent has about another (Pynadath & Marsella, 2005).

Here, as in (Pynadath, Rosenbloom & Marsella 2014), RL is leveraged in selecting among models of other agents; in particular, it is used so that the security agent can learn a model of the intruder. First a form of multiagent RL is used to learn a distinct policy, or Q function, for the intruder under each possible model, and then these policies are used in combination with the perception of the intruder's actions to yield a gradient over the two models that is proportional to the models' Q values for the performed actions. For example, the model for which the observed action has higher Q values will achieve increased likelihood in the posterior distribution. This very quickly enables the security agent to determine the correct door in the experiments run, with the Q value learned by RL substituting for what would otherwise be a need to explicitly extrapolate and compare the paths the intruder might take to the two doors.

The conditional that compares the Q values and generates a posterior distribution for the models is shown in Figure 7. It multiplies the Q values for the observed action – specified by the location of the intruder agent and the direction of movement from that location – in each policy by 0.1 to scale utilities in [0,10] to values for selection in [0,1], and then projects these values onto the model predicate to generate a posterior distribution on the model currently being used by the intruder.

```
CONDITIONAL PREDICT-MODEL
   Conditions: Previous-RL-Loc(location:loc)
               RL-Direction(direction:d)
               Q(model:m location:loc direction:d
                         value:[0.1*q])
   Actions: Model(model:m)
```

**Figure 7.** Conditional for model prediction.

In general, RL enables agents to learn effective policies for task performance based on rewards received over a sequence of trials (Sutton, 1998). In Sigma, RL is not a separate architectural learning algorithm, but occurs through a combination of gradient-descent learning and the appropriate knowledge expressed in predicates and conditionals. This also means that, as in (Pynadath, Rosenbloom and Marsella 2014), it is possible to move from single-agent to multi-agent RL, and from RL given a single reward function to RL given a range of possible reward functions – i.e., models – by appropriately changing the predicates and conditionals.

Combining the short-term adaptivity provided by rule-based behavior with the long-term adaptivity provided by map and ToM learning yields embodied Sigma-based synthetic agents that exhibit effective social and environmental interaction in the physical security model.

### The Appraisal-Based Exploration Model

Emotion research in Sigma is motivated by the hypothesis that emotion is critical for a general intelligence to survive and thrive in complex physical and social worlds. A large fragment of emotion is non-voluntary and immutable, and hence, significant portions of it must be grounded architecturally rather than being primarily activated by learned knowledge and skills (Rosenbloom, Gratch, & Ustun, 2015). Sigma's architectural support for emotional processing leverages *appraisals*, which capture emotional and behavioral-relevant assessments of situations concerning a relatively small set of variables, such as relevance, desirability, likelihood, expectedness, causal attribution, controllability, and changeability in the EMA theory (Marsella & Gratch, 2009).

The primary objective of this model is to demonstrate how architectural support for appraisals can be utilized as a heuristic for an item-location task in a virtual environment, even when there is no algorithmic knowledge about how to explore. In this task, the agent is asked to locate an item while building a map of the environment. The setup is almost identical to the previous model for the physical security system scenario in terms of what is being communicated to the Sigma models, the grid-based representation, and what has been delegated to the gaming engine. The only difference is that the appraisal-based exploration model is realized in Unity rather than the SmartBody animation engine (Figure 8). The Sigma cognitive model for this task leverages a `Map(x:location y:location object:object%)`predicate that includes a function that learns a probability distribution over objects that are located on the grid. This approach utilizes three appraisals – unexpectedness (or surprise), desirability, and familiarity – that are automatically updated by the architecture in the context of the `Map` predicate.

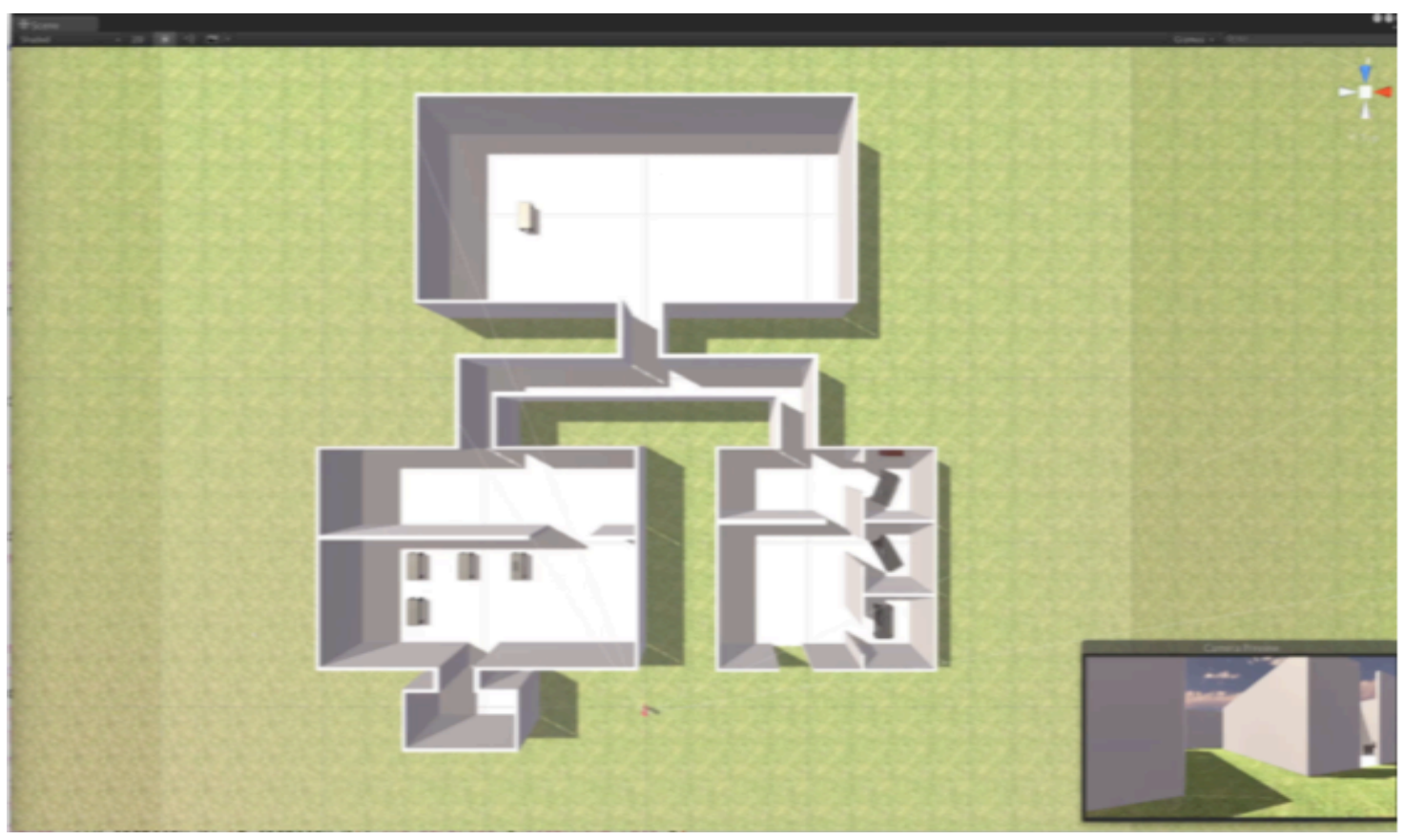

**Figure 8.** Appraisal based exploration in Unity.

Unexpectedness concerns whether an event is not well predicted by existing knowledge, mapping onto the notion of surprise that underlies the bottom-up aspects of today's leading models of visual attention. In other words, attention is drawn to what is surprising or unexpected. The Bayesian Theory of Surprise, which is adapted here, compares the prior distribution over the visual field – i.e., the model that has previously been learned for it – with the posterior distribution derived via Bayesian belief updating of the prior given the image (Itti and Baldi, 2006). The size of the difference correlates with how poorly past knowledge predicts the current event. The computation of surprise in Sigma tracks this approach but differs in several details. Distribution updating is mediated by Sigma's gradient-descent learning, with the functions before and after learning compared (Rosenbloom, Gratch, & Ustun, 2015). For example, any time an object is perceived in the appraisal-based exploration model, the function for the `Map` predicate is updated, yielding an architectural measure for surprise. The more significant the update, the higher the surprise is. For example, when an object is seen for the first time, the corresponding update generates the most surprise for the perceived object unless the perceptions are significantly noisy.

Desirability concerns whether or not an event facilitates or thwarts what is wanted. In Sigma, it is modeled as a relationship between the current state and the goal. Given a predicate's state and goal, desirability amounts to how similar/different the state is to/from the goal, calculated using Hellinger distance (Rosenbloom, Gratch, & Ustun, 2015). For the appraisal-based exploration model, the goal for the `Map` predicate is set as the item for which the search is being conducted, resulting in higher similarity values wherever it is located. Combining surprise and progress for the `Map` predicate via an approximation to *probabilistic or* yields an attention function for the objects in the map. The surprise measure acts as the bottom-up input and the progress (or difference) measure acts as the top-down input to this attention function. In this appraisal-based exploration model, the agent attends (visits) the object with the highest attention measure, which is the sought object if it is currently perceived – with the top-down input dominating here – or an unfamiliar/surprising object otherwise. This latter bottom-up component reflects one key aspect of curiosity, in seeking things that are surprising when what is desired is not available. When there are no unvisited objects in the visual field, the agent leverages the familiarity appraisal to pick a location on the grid to visit. This appraisal simply counts the number of times a region in a learned function – a grid location for the function of the `Map` predicate in this particular case – is updated. When the familiarity measure is triggered, the agent randomly selects one of the least visited areas on the grid, as those areas would be the least updated regions in the function. This captures a second key aspect of curiosity, by driving exploration towards what is less familiar. In the present set up, curiosity about surprising objects trumps curiosity about unknown regions.

In summary, the combination of surprise and desirability, as attention, guide the exploitation process, whereas the combination of familiarity and surprise, as curiosity, guide the exploration process; with no task-specific knowledge required. In the experiments run for this model, the agent always manages to locate the sought item, but as expected, this may take significantly longer than it would if behavior were driven by knowledge-intensive algorithms specifically designed for this task.

## CONCLUSIONS

Work to date on Sigma has explored various capabilities individually, including learning (Rosenbloom, 2012a; Rosenbloom et al., 2013; Rosenbloom 2014), memory and knowledge (Rosenbloom, 2010; Rosenbloom, 2014; Ustun et al., 2014), decision making and problem solving (Chen et al., 2011; Rosenbloom, 2011), perception (Chen et al., 2011), speech (Joshi, Rosenbloom & Ustun 2014; Joshi, Rosenbloom and Ustun, 2016), Theory of Mind (Pynadath et al., 2013; Pynadath, Rosenbloom and Marsella 2014), emotions (Rosenbloom, Gratch & Ustun, 2015), and more recently neural networks (Rosenbloom, Demski & Ustun, 2016b; Rosenbloom, Demski & Ustun, 2017). The models discussed in this paper explore multiple forms of integrative proof-of-concept cognitive models that utilize varying combinations of capabilities as early attempts toward developing integrative HASC systems that can be integral to interactive simulations. Still, this is just the beginning of a long journey. One thing that is required is more of the capabilities necessary for complete HASC systems. Capabilities currently under development include enhanced neural processing, a motivational system, language processing, and additional forms of learning – particularly structure learning.

HASC systems in general require a computational intelligent behavior model that pushes the boundaries of how such a broad range of capabilities may be integrated. Such integration is an essential goal for cognitive architectures, and Sigma provides a unique and potentially powerful way of doing this by leveraging the synergy between diverse capabilities that is facilitated by grand unification and functional elegance, hopefully leading to *plug compatibility* between humans and artificial systems in interactive simulations by generating characters that behave as humanly as possible, that are autonomous, and that are socially interactive through language, *Theory of Mind*, and emotional capabilities. It is our intent to continue pushing both the capabilities available within Sigma and their integration, initially through a diverse range of smaller prototype agents, but ultimately through realistic agents that can be of practical use. One example for prototype agents is our current work towards Sigma models for the RoboCup 2D Soccer Simulation League. There are also plans for extending the basic agents in the physical security model to incorporate natural language, speech, and affect for interactive training simulation scenarios.

## ACKNOWLEDGEMENTS

This effort has been sponsored by the Office of Naval Research and the U.S. Army. Statements and opinions expressed do not necessarily reflect the position or the policy of the United States Government, and no official endorsement should be inferred.